\documentclass[10pt,twocolumn,letterpaper]{article}
\usepackage{multirow}
\usepackage{cvpr}              

\usepackage{graphicx}
\usepackage{amsmath}
\usepackage{amssymb}
\usepackage{booktabs}

\usepackage[T1]{fontenc}
\usepackage[utf8]{inputenc}
\usepackage[accsupp]{axessibility}
\usepackage{color}
\usepackage[dvipsnames]{xcolor}
\usepackage[pagebackref,breaklinks,colorlinks,citecolor=cvprblue]{hyperref}

\definecolor{cvprblue}{rgb}{0.21,0.49,0.74}

\usepackage[capitalize]{cleveref}
\crefname{section}{Sec.}{Secs.}
\Crefname{section}{Section}{Sections}
\Crefname{table}{Table}{Tables}
\crefname{table}{Tab.}{Tabs.}

\def\paperID{736}
\def\confName{CVPR}
\def\confYear{2024}

\title{PLGSLAM: Progressive Neural Scene Represenation with Local to Global Bundle Adjustment }

\author{Tianchen Deng\textsuperscript{1}, Guole Shen\textsuperscript{1}, Tong Qin\textsuperscript{1}, Jianyu Wang\textsuperscript{1}, Wentao Zhao\textsuperscript{1},\\
Jingchuan Wang\textsuperscript{1}, Danwei Wang\textsuperscript{2}, Weidong Chen\textsuperscript{1}
\thanks{ represents corresponding author.} \\
{\textsuperscript{\rm 1} Shanghai Jiao Tong University}
{\textsuperscript{\rm 2} Nanyang Technological University}
}

\begin{document}

\maketitle
\begin{abstract}
Neural implicit scene representations have recently shown encouraging results in dense visual SLAM. However, existing methods produce low-quality scene reconstruction and low-accuracy localization performance when scaling up to large indoor scenes and long sequences. These limitations are mainly due to their single, global radiance field with finite capacity, which does not adapt to large scenarios. Their end-to-end pose networks are also not robust enough with the growth of cumulative errors in large scenes. To this end, we introduce PLGSLAM, a neural visual SLAM system capable of high-fidelity surface reconstruction and robust camera tracking in real-time. To handle large-scale indoor scenes, PLGSLAM proposes a progressive scene representation method which dynamically allocates new local scene representation trained with frames within a local sliding window. This allows us to scale up to larger indoor scenes and improves robustness (even under pose drifts). In local scene representation, PLGSLAM utilizes tri-planes for local high-frequency features with multi-layer perceptron (MLP) networks for the low-frequency feature, achieving smoothness and scene completion in unobserved areas. Moreover, we propose local-to-global bundle adjustment method with a global keyframe database to address the increased pose drifts on long sequences. Experimental results demonstrate that PLGSLAM achieves state-of-the-art scene reconstruction results and tracking performance across various datasets and scenarios (both in small and large-scale indoor environments). The code is open-sourced at \href{https://github.com/dtc111111/plgslam}{https://github.com/dtc111111/plgslam}.
\end{abstract}
\vspace{-0.4cm}
\section{Introduction}
\vspace{-0.2cm}
\label{sec:intro}
Visual Simultaneous Localization and Mapping
(SLAM) has been a fundamental computer vision problem with wide applications such as autonomous driving, remote sensing~\cite{yang1}, and virtual/augmented reality. Many traditional methods have been introduced in the past years, such as ORB-SLAM~\cite{orbslam2,orbslam}, VINS~\cite{vins}, and so on. They can estimate the camera pose and construct sparse point cloud maps in real-time with accurate localization performance. However, the sparse point cloud maps cannot meet the further perception needs of the robot. Recent attention has turned to learning-based methods for dense scene reconstruction. Kinectfusion \cite{kinectfusion}, BAD-SLAM\cite{BADslam}
reconstruct meaningful global 3D maps and show
reasonable but limited reconstruction accuracy with deep learning networks.

Nowadays, with the proposal of Neural Radiance Fields (NeRF), there are many following works on different areas~\cite{prosgnerf}. iMAP~\cite{imap} is the first work to use a single multi-layer perceptron (MLP) to represent the entire scene in SLAM system. NICE-SLAM~\cite{niceslam} improves the scene representation method with feature grids.
ESLAM~\cite{eslam} and Co-SLAM~\cite{coslam} further improve the scene representation methods. ESLAM uses tri-planes for better real-time performance and reconstruction accuracy. Co-SLAM uses joint coordinate and sparse parametric scene for accurate scene representation. They can achieve promising reconstruction quality in a small indoor room.

Although ESLAM and Co-SLAM perform well in smaller indoor scenes, they face challenges in representing large-scale indoor scenes (e.g., multi-room apartments). We outline the key challenges for real-time incremental NeRF-SLAM: \textit{\textbf{a) insufficient scene representation capability:}} Existing methods employ a fixed-capacity, global model, limiting scalability to larger scenes and longer video sequences. \textit{\textbf{b) accumulation of errors and pose drift:}} Existing works struggle with accuracy and robustness in large-scale indoor scenes due to accumulating errors.    

To this end, we design our neural SLAM system for accurate scene reconstruction and robust pose estimation in large indoor scenes and long sequences. We propose a progressive scene representation method which dynamically initialize new scene representation when the camera moves to the bound of the local scene representation. The entire scene is divided into multiple local scene presentations, which can significantly improve the scene representation capacity of large indoor scenes. The robustness of our system is also increased because the mis-estimation is locally bounded. 

In local scene representation, We propose a parametric-coordinate joint encoding method for accuracy, speed, and completion of unseen region. Parametric encoding is the tri-plane encoding, and the coordinate encoding is the one-blob encoding with MLP. We use tri-planes to encode the local high-frequency feature of the scene and use MLP to represent global low-frequency features with the coherence priors inherent. We bring together the benefits of both methods for accuracy, smoothness, and hole-filling in areas without observation.   

Furthermore, we combine the traditional SLAM systems with end-to-end pose networks to improve pose estimation performance. We propose a local-to-global bundle adjustment (BA) method to eliminate the cumulative error which becomes significantly evident in large indoor scenes and long video sequences. So far, all neural SLAM systems only use end-to-end network and perform BA with rays sampled from a local subset of selected keyframes, resulting in inaccurate, non-robust pose estimation and significant cumulative errors in camera tracking. PLGSLAM maintains a global keyframe database and performs local-to-global neural warpping and reprojection Bundle Adjustment. The proposed Local-to-global BA method can eliminate the cumulative error with all the historical observations. In practice, PLGSLAM achieves SOTA performance in camera tracking and
3D reconstruction while maintaining real-time performance. \textbf{Overall, our contributions are shown as follows:}
\begin{itemize}
    \item A progressive scene representation method is proposed which dynamically initiate local scene representation trained with frames within a local window. This enables scalability to extensive indoor scenes and long videos sequences, substantially improving robustness.

    \item  In local scene representation, We design a joint parametric-coordinate encoding method. We combine the tri-planes with the one-blob encoding encoding method for accurate and smooth surface reconstruction. It can not only enhance the ability of scene representation, but also substantially reduce the memory growth from cubic to square.

    \item  We integrate the traditional SLAM system with an end-to-end pose estimation network. A local-to-global bundle adjustment algorithm is proposed, which can mitigate cumulative error in large-scale indoor scenes. Our system maintain a global keyframe database with the system operation, enabling bundle adjustment across all past observations, from local to global.

\end{itemize}
\begin{figure*}[t]
  \centering
   \includegraphics[width=\linewidth]{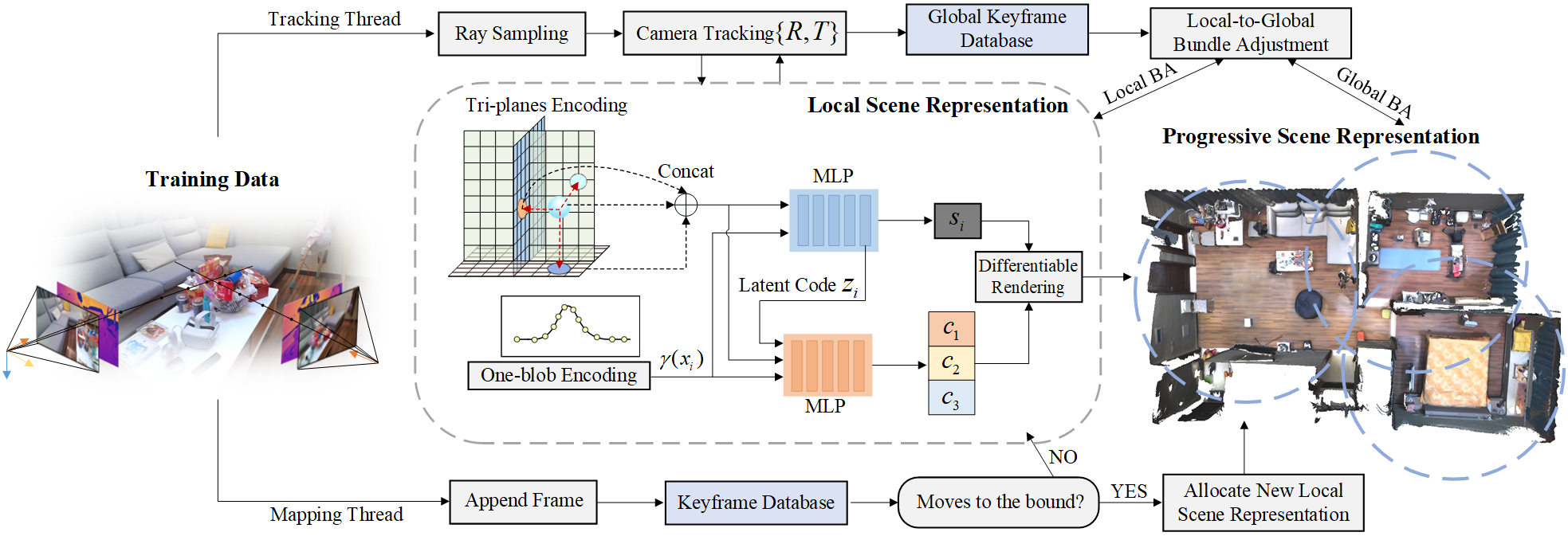}
   \vspace{-0.4cm}
    \caption{The isometric view of the proposed PLGSLAM system. Our system has two parallel threads: the mapping thread and the tracking thread. In the mapping thread, we propose the progressive scene representation method for the entire scene. In local scene representation, we combine the tri-planes with the multi-layer perceptron to improve the accuracy and smoothness. Both of them are online updated by minimizing our carefully designed loss through differentiable rendering with the system operating. As for the tracking thread, we propose a local-to-global bundle adjustment for accurate and robust pose estimation. Those two threads are running with an alternating optimization. }
   \label{fig:pipeline}
   \vspace{-0.6cm}
\end{figure*}
\vspace{-0.2cm}
\section{Related Work}
\vspace{-0.2cm}
\noindent\textbf{Dense Visual SLAM}.\quad SLAM~\cite{liu1} and localization~\cite{liu2,zhang1,zhang2,peng1} has been an active field for the past two decades. Traditional visual SLAM algorithms~\cite{orbslam,vins,deng,xie,xie2} estimate accurate camera poses and use sparse point
clouds as the map representation. They use manipulated key points for tracking, mapping, relocalization, and loop closing. Dense visual SLAM approaches focus on reconstructing a dense map of a scene. DTAM \cite{dtam} is one of the pioneer works that use the dense map and view-centric scene representation. KinectFusion~\cite{kinectfusion} performs camera tracking via projective iterative-closest-point (ICP)
and explicitly represents the surface of the environment via TSDF-Fusion. Some works \cite{BADslam,Bundle-fusion, ba-net} propose bundle adjustment(BA) method to optimize keyframe poses and construct the dense 3D structure jointly.   In contrast to
previous SLAM approaches, we adopt implicit scene representation of the geometry and directly optimize them during mapping.

\noindent \textbf{Implicit Scene Representation.} \quad With the proposal of Neural radiance fields (NeRF)~\cite{NeRF}, many researchers explore taking the advantages of the implicit method into 3D reconstruction. NeRF is a ground-breaking method for novel view synthesis using differentiable rendering. However, the representation of volume densities can not commit the geometric consistency. In order to deal with it, UNISURF~\cite{UNISURF} and NeuS~\cite{NEUS} are proposed, combining neural radiance fields with Signed Distance Field (SDF) values. Other methods~\cite{neural_rgbd, transformerfusion, VolumeFusion, neuralrecon} use various scene geometry representation methods, such as truncated signed distance function, voxel grid. For large-scale representation, Mega-NeRF and LocalRF~\cite{meganerf, progressive} use multiple local scene representations for the entire scene.

\noindent \textbf{NeRF-based SLAM.} \quad
iMAP~\cite{imap} and NICE-SLAM~\cite{niceslam} are successively proposed to combine neural implicit mapping with SLAM. iMAP uses a single multi-layer perceptron (MLP) to represent the scene, and NICE-SLAM uses a learnable hierarchical feature grid. \cite{zhu1,li1,li2} use semantic feature embedding to improve scene representation. Some methods~\cite{compact,zhu2} also use 3D gaussian to improve the scene representation. Co-SLAM~\cite{coslam} and ESLAM~\cite{eslam} are the most relative work of our method. However, all of them have difficult in large-scale indoor environments and long sequences. With the proposed progressive scene representation method, we can successfully scale up to larger indoor scenarios. The fusion of tri-planes and one-blob encoding leads to high-fidelity and smooth surface reconstruction in local scene. A local-to-global bundle adjustment method is also proposed. This method can effectively eliminate growing cumulative errors in existing methods in large indoor scenes. 
\vspace{-0.2cm}
\section{Method}
\vspace{-0.2cm}
The pipeline of our system is shown in Fig.~\ref{fig:pipeline}. We use a set of sequential RGB-D frames $\{I_i, D_i\}_{i=1}^M$ with known camera intrinsic $K \in R_{
3\times3}$ as our input. Our model predicts camera poses $\{R_i|t_i\}^M_{i=1}$, color $\mathbf{c}$, and an implicit truncated signed distance function (TSDF) representation $\phi_{\boldsymbol{g}}$ that can be used in marching cubes algorithm to extract 3D meshes. For the implicit mapping thread, a progressive scene representation method (Sec.~\ref{sec:progressive}) is designed to represent large-scale indoor environments. Then, in the local radiance fields, we improve the scene representation methods and combine the  tri-planes with multi-layer perceptron (MLP) by our designed architecture. Sec.~\ref{sec:render}
walks through the rendering process, which converts raw representations into pixel depths, colors, and SDF values. For the camera tracking thread, a local-to-global bundle adjustment method (Sec.~\ref{sec:localtoglobal}) is designed for robust and accurate pose estimation. Several carefully designed loss functions are proposed to jointly optimize the scene implicit representation and camera pose estimation. The network is incrementally updated with the system operation. 
\vspace{-0.2cm}
\subsection{Progressive Scene Representation}

\label{sec:progressive}
All the existing NeRF-based SLAM systems have difficulties in large-scale indoor scenes. They use a single, global representation of the entire environment, which limits their scene representation capacity. 
There are two key limitations when modeling large-scale indoor scenes: \textit{\textbf{a)} the incapacity of a single, fixed-capacity model to represent videos of arbitrary length. \textbf{b)} the single scene representation tends to overfit to the early data in the sequence, leading to poorer performance in learning from the later data. \textbf{c)} any mis-estimation (e.g. outlier pose) has a global impact and might cause the false reconstruction.  }  

Mega-NeRF and Bungee-NeRF~\cite{meganerf,bungee} pre-partition the space for radiance fields. However, this approach is not applicable in our setting, as the camera poses in our system are concurrently optimized alongside the mapping thread.

 In our method, we dynamically create local scene representation. Whenever the estimated camera pose trajectory leaves the space of the current scene representation, we dynamically allocate new local scene representation trained with a small set of frames. and from there, we progressively introduce subsequent local frames to the optimization. So, the entire scene can be represented as multiple local scene representations: 
\begin{equation}
\{I_i,D_i\}_{i=1}^M\mapsto \{\mathrm{SR}^1_{\theta_1},\mathrm{SR}^2_{\theta_2},\dots ,\mathrm{SR}^n_{\theta_n}\}\mapsto\{\mathbf{c},\mathbf{\sigma}\}
\end{equation}
where $SR_{\theta_n}^n$ denotes the local scene representation, $\sigma$ denotes the volume density. Each local scene representation is centered at the position of the last estimated camera pose. 
We train each scene representation with a local subset frames. Each subset contains some overlap frames, which is important for achieving consistent reconstructions in the local scene representation. Whenever the estimated camera pose leaves the bound of the current scene representation, we stop optimizing previous ones (freeze the network parameters). At this point, we can reduce memory requirements by removing unnecessary supervisory frames.
We also stop updating the mapping parameters in the tracking thread to reduce errors. If the estimated camera pose is outside the current bounds, but within a previous local scene representation, we activate the previous one and proceed with the optimization process. We further increase the global consistency by inverse distance weight (IDW) fusion for all overlapping scene representations at any supervising frame. 

\noindent\textbf{Local Scene Representation.}
Voxel grid-based architectures~\cite{plenoxels,niceslam,coslam} are the mainstream in NeRF-based SLAM system. However, they struggle with cubical memory growing and real-time performance. Inspired by \cite{eslam}, we design a parametric-coordinate joint encoding method. Parametric encoding is tri-plane encoding, and the coordinate encoding is the one-blob encoding with MLP. We store and optimize high-frequency features(e.g. texture) on perpendicular axis-aligned planes. The one-blob encoding with MLPs are used to encode and store low-frequency features for the coherence and smoothness priors. This joint scene representation architecture achieve high-fidelity and smoothness scene reconstruction with the ability of hole filling. 

Specifically, the tri-planes are at two scales, i.e., coarse and fine. The tri-planes feature $\boldsymbol{T}(x)$ can be formulated as:

\begin{align}
t^c(x) & =T_{x y}^c(x)+T_{x z}^c(x)+T_{y z}^c(x)\notag \\
t^f(x) & =T_{x y}^f(x)+T_{x z}^f(x)+T_{y z}^f(x) \notag\\
\boldsymbol{T}(x) & =Concat\left(t^c(x) ; t^f(x)\right)
\end{align}
where $t^c(x),t^f(x)$ denote the coarse and fine feature form tri-planes. $x$ is the world coordinate. $\{T^c_{x y},T^c_{x z},T^c_{y z}\}$ represent the three coarse geometry feature planes, and  $\{T^f_{x y},T^f_{x z},T^f_{y z}\}$ represent the three fine geometry feature planes.

For a sample point $x$, we use bilinearly interpolating the nearest neighbors on each feature plane. Then, we sum the interpolated coarse features and the fine, respectively, into the coarse output and fine output. At last, we concatenate the outputs together as the tri-plane features.
 The geometry decoder outputs the predicted SDF value $\phi_{ \boldsymbol{g}}(x)$ and a feature vector $\mathbf{z}$:
\begin{equation}
f_g\left(\gamma(x),  \boldsymbol{T}(x)\right) \to(\mathbf{z}, \phi_{ \boldsymbol{g}}(x))
\end{equation}
where $\gamma(\mathbf{x})$ represents coordinate position encoding. $\mathbf{z}$ is the latent code. We use one-blob  encoding~\cite{coslam} instead of embedding spatial coordinates into multiple frequency bands. Finally, the color decoder predicts the RGB value:
\begin{equation}
f_c(\gamma(x), \mathbf{z}, \boldsymbol{a}(x)) \mapsto \phi_{\boldsymbol{a}}(x)
\end{equation}
$\phi_{\boldsymbol{a}}(x)$ represents the color of the sample points. $\boldsymbol{a}(x)$ is the is the appearence feature from tri-planes. Combining the MLP
with the tri-planes scene representation, our architecture achieve accurate and smooth surface reconstruction, efficient memory use, and hole filling performance.

\vspace{-0.2cm}
\subsection{Differentiable Rendering}
\vspace{-0.2cm}
\label{sec:render}
 Inspired by the recent success of volume rendering in NeRF~\cite{NeRF}, we also propose to use a differentiable rendering process to integrate the predicted density and colors from our scene representation. We determine a ray $r(t)=\mathbf{o}+t\mathbf{d}$ whose origin is at the camera center of projection $o$, ray direction $r$. We uniformly sample K points. The sample bound is within the near and far planes $t_k\in [t_n,t_f]$, $k \in \{1,\dots,K\}$ with depth values $\{\mathbf{d_1}, \dots, \mathbf{d_K}\}$ and predicted colors $\{\mathbf{c_1}, . . . , \mathbf{c_K}\}$. For all sample points along rays, we query TSDF $\phi_{ \boldsymbol{g}}(p_k)$ and raw color $\phi_{\boldsymbol{a}}(p_k)$ from our
networks and use the SDF-Based rendering approach to convert SDF values to volume densities:
\begin{equation}
    \boldsymbol{\sigma}\left(x_k\right)=\frac{1}{\beta} \cdot \operatorname{Sigmoid}\left(\frac{-\phi_{\boldsymbol{g}}\left(x_k\right)}{\beta}\right)
\end{equation}
where $\beta \in \mathbb{R} $ is a learnable parameter that controls the sharpness of the surface boundary.
Then we define the termination probability $w_k$, depth $\hat{\boldsymbol{d}}$, and color $\hat{\boldsymbol{c}}$ as:
\begin{equation}
\begin{gathered}
w_k=\exp \left(-\sum_{m=1}^{n-1} \boldsymbol{\sigma}\left(x_m\right)\right)\left(1-\exp \left(-\boldsymbol{\sigma}\left(x_k\right)\right)\right) \\
\hat{\boldsymbol{c}}=\sum_{k=1}^N w_k \boldsymbol{\phi}_{\boldsymbol{a}}\left(x_k\right) \quad \text { and } \quad \hat{\boldsymbol{d}}=\sum_{k=1}^N w_k t_k
\end{gathered}
\end{equation}

\subsection{Local-to-global Bundle Adjustment}
\label{sec:localtoglobal}
\begin{figure}[t]
  \centering
   \includegraphics[width=\linewidth]{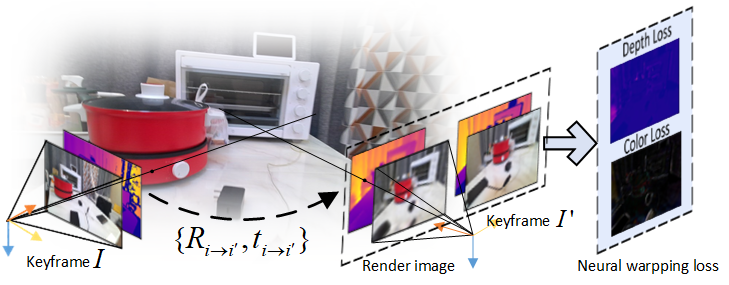}
    \caption{This figure illustrates the designed neural warping loss. We calculate the neural warpping loss between keyframe $I$ and keyframe $I'$.}
   \label{fig:warpping}
   \vspace{-0.4cm}
\end{figure}

Currently, existing NeRF-based SLAM methods exhibit poor accuracy in large-scale indoor scene localization. Their tracking networks are performed via minimizing rgb loss functions with
respect to learnable parameters $\theta$ to estimate the relative pose matrix $\{R_i |t_i \} \in \mathbb{S E}(3)$.
With the growing cumulative error $ \varepsilon $ of pose estimation, those methods result in failure in large-scale indoor scenes and long videos.
To this end, we design a local-to-global bundle adjustment method to solve this problem, which performs well with our progressive scene representation. We design our method
by drawing inspiration from traditional keyframe-based SLAM systems for improving
the robustness and accuracy of pose estimation.
 We propose neural warpping error and reprojection error for local-to-global bundle adjustment. The neural warpping loss is formulated as:
 \vspace{-0.1cm}
\begin{align}
\vspace{-0.4cm}
   \mathcal{L} _{nwc}=\sum_i^N(\mathcal{F}( o_i,d_i,\{R_{i\to i'},t_{i\to i' }\} )-\boldsymbol{C}_{i'}) \notag \\ 
   \mathcal{L} _{nwd}=\sum_i^N(\mathcal{F}( o_i,d_i,\{R_{i\to i'},t_{i\to i' }\} )-\boldsymbol{D}_{i'}) 
   \vspace{-0.4cm}
\end{align}
\vspace{-0.1cm}
Here, $L_{nwc}$ and $L_{nwd}$ are the neural warpping color and depth loss. $o_i,d_i$ denotes the rays from image $I_i$. $\{R_{i\to i'},t_{i\to i' }\}$ denotes the relative pose from image $I_i$ to $I_{i'}$. $\mathcal{F}()$ denotes our scene representation network. We present the illustration of neural warpping loss in Fig. \ref{fig:warpping}. We formulate reprojection errors with SIFT features:
\vspace{-0.1cm}
\begin{equation}
\vspace{-0.2cm}
    \mathcal{L}_{re} =\sum_{i=1}^n\left \| (u_{i'},v_{i'})-\Pi(R_{i\to i'}P_i+t_{i\to i' })   \right \| 
    \vspace{-0.2cm}
\end{equation}

where $\Pi(R_{i\to i' }P_i+t_{i\to i' })$ represents the reprojection of 3D point $P_i$ to the corresponding pixel $(u_{i'},v_{i'})$ in image $i'$.  

Whenever a keyframe arrives, we perform local bundle adjustment in our tracking and mapping thread. A keyframe is selected for every $K$ frames. When the camera moves to the bound of the current scene representation, we also initialize it as the keyframe. In local bundle adjustment, we only select keyframes from the local keyframe database that visually overlap with the current frame when optimizing the scene geometry to ensure the geometry outside the current view remains static and fast convergence. Meanwhile, we also maintain a global keyframe list with the operation of our system. After accumulating a specific number of local keyframes or the camera moves to the local bound, a global bundle adjustment is performed. In global BA, we randomly select keyframes and rays from the global keyframe database, which leverages all historical observations of the scene. This approach effectively integrates local and global information which greatly improves the accuracy of camera pose optimization in large-scale indoor scenes.  
\begin{figure*}[t]
  \centering
   \includegraphics[width=\linewidth]{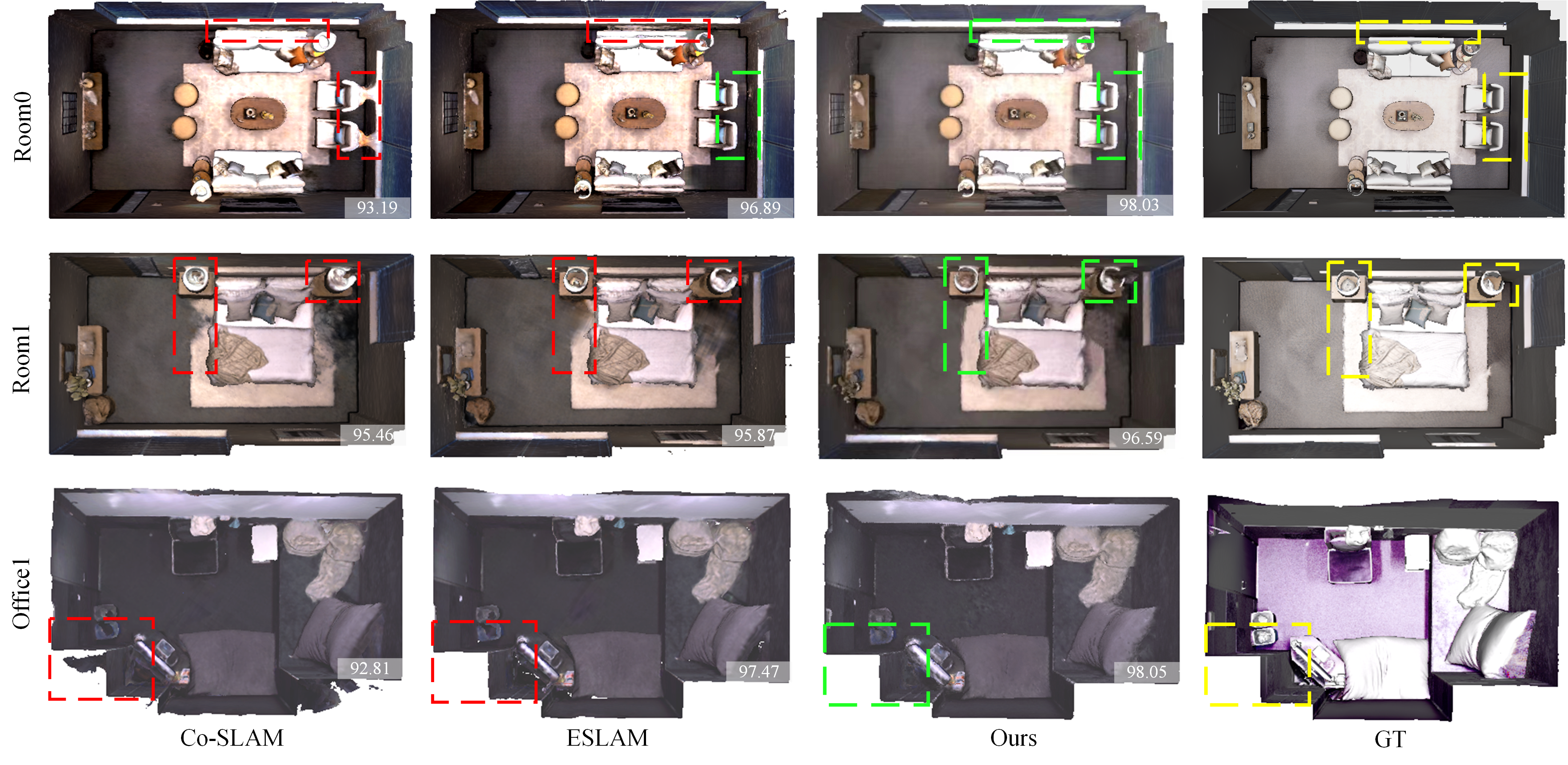}
    \vspace{-0.5cm}
    \caption{ Reconstruction results (without cull) on Replica \cite{replica} apartment dataset. In comparison to our baselines, our methods achieve accurate and high-quality scene reconstruction and completion on various scenes.The region outlined on the image is marked in red to signify lower predictive accuracy, in green to signify higher accuracy, and in yellow to represent the ground truth results. The number in the bottom right corner of the image represents the completion ratio metric.}
    \vspace{-0.3cm}
   \label{fig:replica}
\end{figure*}

\begin{table*}[]
\centering
\scalebox{0.95}{
\setlength{\tabcolsep}{0.9mm}{
\begin{tabular}{l|cccc|cc}
\toprule
\multirow{2}{*}{Methods} & \multicolumn{4}{c|}{Reconstruction}                                  & \multicolumn{2}{c}{Localization}    \\
                         & Depth L1{[}cm{]} $\downarrow$ & Acc.{[}cm{]} $\downarrow$   & Comp.{[}cm{]} $\downarrow$  & Comp.Ratio(\%) $\uparrow$  & ATE Mean{[}cm{]} $\downarrow$ & ATE RMSE{[}cm{]} $\downarrow$ \\ \midrule
iMAP \cite{imap}                     & 4.645            & 3.624          & 4.934          & 80.515          & 3.118            & 4.153            \\
NICE-SLAM \cite{niceslam}             & 1.903            & 2.373          & 2.645          & 91.137          & 1.795            & 2.503            \\
Vox-Fusion \cite{voxfusion}             & 2.913            & 1.882          & 2.563          & 90.936          & 1.067            & 1.453            \\
ESLAM \cite{eslam}                  & 0.945            & 2.082          & 1.754          & 96.427          & 0.565            & 0.707            \\
Co-SLAM \cite{coslam}                & 1.513            & 2.104          & 2.082          & 93.435          & 0.935            & 1.059            \\
Ours              & \textbf{0.771}   & \textbf{1.793} & \textbf{1.543} & \textbf{97.877} & \textbf{0.525}   & \textbf{0.635}   \\  \bottomrule
\end{tabular}}}
\vspace{-0.2cm}
\caption{Quantitative results of our proposed PLGSLAM with existing NeRF-based SLAM system on the Replica dataset \cite{replica}. We evaluate reconstruction and localization performance in small room scenes. The results are the average on the scenes of the Replica dataset. Our method outperforms the existing method in surface reconstruction and pose estimation.  }
\vspace{-0.4cm}
\label{tab:replica}
\end{table*}
\begin{table}[]
\centering
\scalebox{0.85}{
\setlength{\tabcolsep}{0.8mm}{
\begin{tabular}{l|ccc|cc}
\hline
\multirow{2}{*}{Methods} & \multicolumn{3}{c|}{Reconstruction{[}cm{]}}           & \multicolumn{2}{c}{Localization{[}cm{]}} \\
                            & Acc.       & Comp.      & Comp.Ratio(\%) & Mean            & RMSE           \\ \hline
NICE-SLAM\cite{niceslam} & 29.17              & 4.45          & 67.97          & 8.78             & 9.63          \\
ESLAM\cite{eslam}     & 26.22              & 4.53          & 71.43          & 7.89             & 8.95          \\
Co-SLAM\cite{coslam}   & 26.55              & 4.67          & 70.34          & 7.67             & 8.75          \\
Ours      & \textbf{19.42}     & \textbf{4.21} & \textbf{74.48} & \textbf{6.12}    & \textbf{6.77}       \\ \hline
\end{tabular}}}
\caption{Camera tracking results on the Scannet datasets~\cite{scannet}.  We evaluate our camera tracking performance on the Scannet dataset to verify the effectiveness of our method. Our method achieves high-fidelity surface reconstructions and superior camera tracking.}
\vspace{-0.3cm}
\label{tab:scannet}
\vspace{-0.5cm}
\end{table}
\vspace{-0.2cm}
\subsection{Objective Functions}
\vspace{-0.2cm}
Our mapping and tracking thread are performed via minimizing our objective functions with respect to network parameters $\theta$ and camera parameters $\{R_i|t_i\}$.
The color and depth rendering losses are used in our mapping and tracking thread:
\begin{equation}
\mathcal{L}_{c}=\frac{1}{N} \sum_{i=1}^N\left(\hat{\mathbf{c}}_i-\mathbf{C_i}\right)^2,\quad
\mathcal{L}_d=\frac{1}{\left|R_i\right|} \sum_{i \in R_i}\left(\mathbf{\hat{d_i}}-\mathbf{D_i}\right)^2
\end{equation}
\vspace{-0.1cm}
where $R_i$ is the set of rays that have a valid depth observation. In addition, we design SDF loss, free space loss, and feature smoothness losses for our mapping thread. Specifically, for samples within the truncation region, we leverage the depth sensor measurement to approximate the signed distance field:
\begin{equation}
    \mathcal{L}_{s d f}=\frac{1}{\left|R_i\right|} \sum_{r \in R_i} \frac{1}{\left|X_r^{t r}\right|} \sum_{x \in X_r^{t r}}\left(\phi_{\boldsymbol{g}}(x)\cdot T-(\mathbf{D_i}-\mathbf{d})\right)^2
\end{equation}
where $X_r^{t r}$ is a set of points on the ray r that lie in the truncation region, $|\mathbf{D_i} - \mathbf{d}| \le  tr$. We differentiate the weights of points that are closer
to the surface $X_r^{tm}=\{x|x\in|\mathbf{D_i} - \mathbf{d}| \le  0.4tr\}$ from those that are at the tail of the truncation region $X_r^{tt}$  in our SDF loss. 
\begin{equation}
    \mathcal{L}_{sdf_m}=\mathcal{L}_{sdf}\left(X_r^{tm}\right), \quad  \mathcal{L}_{sdf_t}=\mathcal{L}_{sdf}\left(X_r^{tt}\right)
\end{equation}
For sample points that are far from the surface $|D_i - d| \ge  T$: 
\begin{equation}
    \mathcal{L}_{f s}=\frac{1}{|R_i|} \sum_{r \in R_i} \frac{1}{\left|X_r^{f s}\right|} \sum_{x \in X_r^{f s}}\left(\phi_{\boldsymbol{g}}(x)-1\right)^2
\end{equation}
This loss can force the SDF prediction value to be the truncated distance tr. In addition, we propose feature smoothness losses to prevent the noisy reconstructions caused by tri-planes in unobserved free-space regions:
\begin{equation}
\mathcal{L}_{\text {smooth }}=\frac{1}{|\mathcal{M}|} \sum_{\mathbf{x} \in \mathcal{M}} \Delta_{xy}^2+\Delta_{xz}^2+\Delta_{yz}^2
\end{equation}
where $\Delta_{xy}=\mathbf{T}\left(\mathbf{x}+\epsilon_{x,y}\right)-\mathbf{T}(\mathbf{x})$, $\Delta_{xz}=\mathbf{T}\left(\mathbf{x}+\epsilon_{x,z}\right)-\mathbf{T}(\mathbf{x})$, $\Delta_{yz}=\mathbf{T}\left(\mathbf{x}+\epsilon_{y,z}\right)-\mathbf{T}(\mathbf{x})$ denotes the feature-metric difference between adjacent sampled vertices on the three feature planes. $\mathcal{M}$ denotes a small random region form tri-planes. This loss can enhance the smoothness of our surface reconstruction results and it is only used in mapping thread.

\begin{table*}[]
\centering
\scalebox{0.95}{
\setlength{\tabcolsep}{0.8mm}{
\begin{tabular}{l|cccc|cc}
\toprule
\multirow{2}{*}{Methods} & \multicolumn{4}{c|}{Reconstruction}                                  & \multicolumn{2}{c}{Localization}    \\
                         & Depth L1{[}cm{]} $\downarrow$ & Acc.{[}cm{]} $\downarrow$   & Comp.{[}cm{]} $\downarrow$  & Comp.Ratio(\%) $\uparrow$  & ATE Mean{[}cm{]} $\downarrow$ & ATE RMSE{[}cm{]} $\downarrow$ \\ \midrule
iMAP \cite{imap}                    & 24.558                 & 14.296             & 7.476              &  44.422              &  9.963                & 10.612              \\
NICE-SLAM \cite{niceslam}               & 37.052          & 6.064       & 5.576        & 71.792        & 4.776           & 5.394           \\
Vox-Fusion \cite{voxfusion}              &43.077                  &26.375             & 9.454              &49.554                &11.473                 &12.754                 \\
ESLAM \cite{eslam}                   & 16.355          & 17.546      & 4.301        & 71.626        & 6.637           & 7.283           \\
Co-SLAM \cite{coslam}                 & 6.702           & 13.355      & 3.666        & 80.486        & 6.182           & 6.891           \\
Ours              & \textbf{6.033}          & \textbf{11.086}      & \textbf{3.261}        & \textbf{85.357}        & \textbf{5.574}           & \textbf{6.228}           \\ \bottomrule
\end{tabular}}}
\vspace{-0.3cm}
\caption{Quantitative results of our proposed PLGSLAM with existing NeRF-based SLAM system on the Apartment dataset \cite{niceslam}. We evaluate reconstruction and localization performance in large-scale multi-room scenes. The results are the average of three runs. Our method outperforms the existing method in surface reconstruction and pose estimation.  }
\label{tab:apartment}
\vspace{-0.3cm}
\end{table*}

\begin{table}[]
\centering
\scalebox{0.85}{
\setlength{\tabcolsep}{0.9mm}{
\begin{tabular}{l|l|c|c}
\toprule
& Method & Speed FPT(s) & Memory Grow.R. \\ \midrule
\multirow{4}{*}{Replica\cite{replica}} & NICE-SLAM\cite{niceslam} & 2.10 & $O(L^3)$ \\
& ESLAM\cite{eslam} & 0.18 & $O(L^2)$ \\
& Co-SLAM\cite{coslam} & 0.16 & $O(L^3)$\\
& Ours & \textbf{0.14} & $O(L^2)$ \\ \midrule
\multirow{4}{*}{Scannet\cite{scannet}} & NICE-SLAM\cite{niceslam} & 3.35 & $O(L^3)$ \\
& ESLAM\cite{eslam} & 0.55 & $O(L^2)$ \\
& Co-SLAM\cite{coslam} & 0.38 &$O(L^3)$ \\
& Ours &\textbf{0.37} & $O(L^2)$\\ \bottomrule
\end{tabular}}}
\vspace{-0.3cm}
\caption{Runtime analysis of our method in comparison with existing ones in terms of average frame processing time (AFPT), and model size growth rate w.r.t. scene side length L. We evaluate these method on replica dataset \cite{replica} and Scannet dataset \cite{scannet}. Our method is greatly faster and the model size grow is significantly reduced from cubic to square.}
\vspace{-0.5mm}
\label{tab:runtime}
\end{table}

\begin{table}[]
\centering
\scalebox{0.85}{
\setlength{\tabcolsep}{0.8mm}{
\begin{tabular}{l|ccc|cc} 
\toprule
\multirow{2}{*}{Methods} & \multicolumn{3}{c|}{Reconstruction[cm]} & \multicolumn{2}{c}{Localization[cm]} \\
 & Acc. & Comp. & Comp.Ratio(\%) & Mean & RMSE \\ 
\midrule
w/o joint enc. & 13.314 & 4.687 & 81.347 & 5.935 & 6.787 \\
w/o prog. & 12.754 & 4.231 & 83.156 & 5.875 & 6.693 \\
w/o lg BA & 12.435 & 4.181 & 83.473 & 5.874 & 6.591 \\
Ours & \textbf{\textbf{11.086}} & \textbf{3.261} & \textbf{85.357} & \textbf{5.574} & \textbf{6.228} \\
\bottomrule
\end{tabular}}}
\vspace{-0.6mm}
\caption{ Ablation study. We conduct experiments on Apartment dataset~\cite{niceslam} to verify the effectiveness of our method. Our full model achieves better
completion reconstructions and more accurate pose estimation results.}
\vspace{-0.8mm}
\label{tab:ablation}
\end{table}
\begin{figure*}[t]
  \centering
   \includegraphics[width=\linewidth]{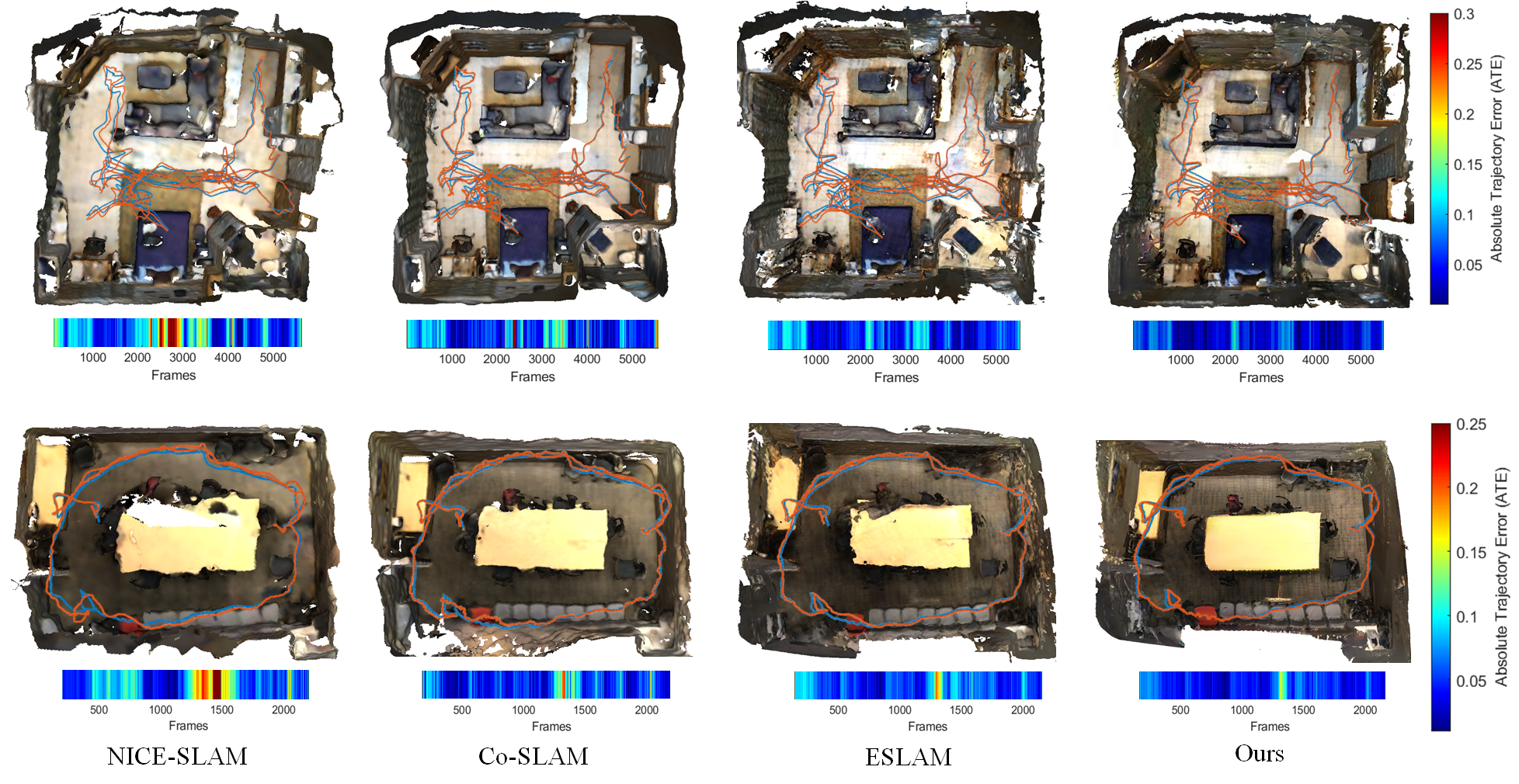}
    \caption{  Qualitative comparison of our proposed PLGSLAM method’s surface reconstruction and localization accuracy with existing NeRF-based dense visual SLAM
methods, NICE-SLAM \cite{niceslam}, Co-SLAM \cite{coslam}, and ESLAM \cite{eslam} on the ScanNet dataset \cite{scannet}. The ground truth camera trajectory is shown in blue, and the estimated trajectory is shown in red. Our method predicts more accurate camera trajectories and does not suffer from drifting issues. We also visualize the Absolute Trajectory Error ATE (bottom color bar) of different methods. The color bar on the right shows the relative scaling of color. It should also be noted that our method runs faster on this dataset.}
\vspace{-0.4cm}
   \label{fig:scannet}
\end{figure*}
\vspace{-0.2cm}
\section{Experiments}
\vspace{-0.2cm}
We validate that our method outperforms existing implicit representation-based methods in surface reconstruction, pose estimation, and real-time performance.
\vspace{-0.2cm}
\subsection{Datasets and Metrics}
\vspace{-0.2cm}
\noindent\textbf{Datasets.} We evaluate PLGSLAM on a variety of scenes from different datasets. We quantitatively evaluate the reconstruction quality on 8 small room scenes from Replica \cite{replica}  (\textbf{nearly}  $6.5m\times4.2m\times2.7m $ with 2000 images). We evaluate on real-world scenes from ScanNet~\cite{scannet} for long sequences (more than 5000 images) and large-scale indoor scenarios (\textbf{nearly} $7.5m\times6.6m\times3.5m $). We also evaluate on Apartment dataset of the multi-rooms scene (\textbf{nearly} $14.5m\times7.5m\times3.8m $ with more than 12000 images) from NICE-SLAM~\cite{niceslam}. 
\begin{figure}[t]
  \centering
   \includegraphics[width=\linewidth]{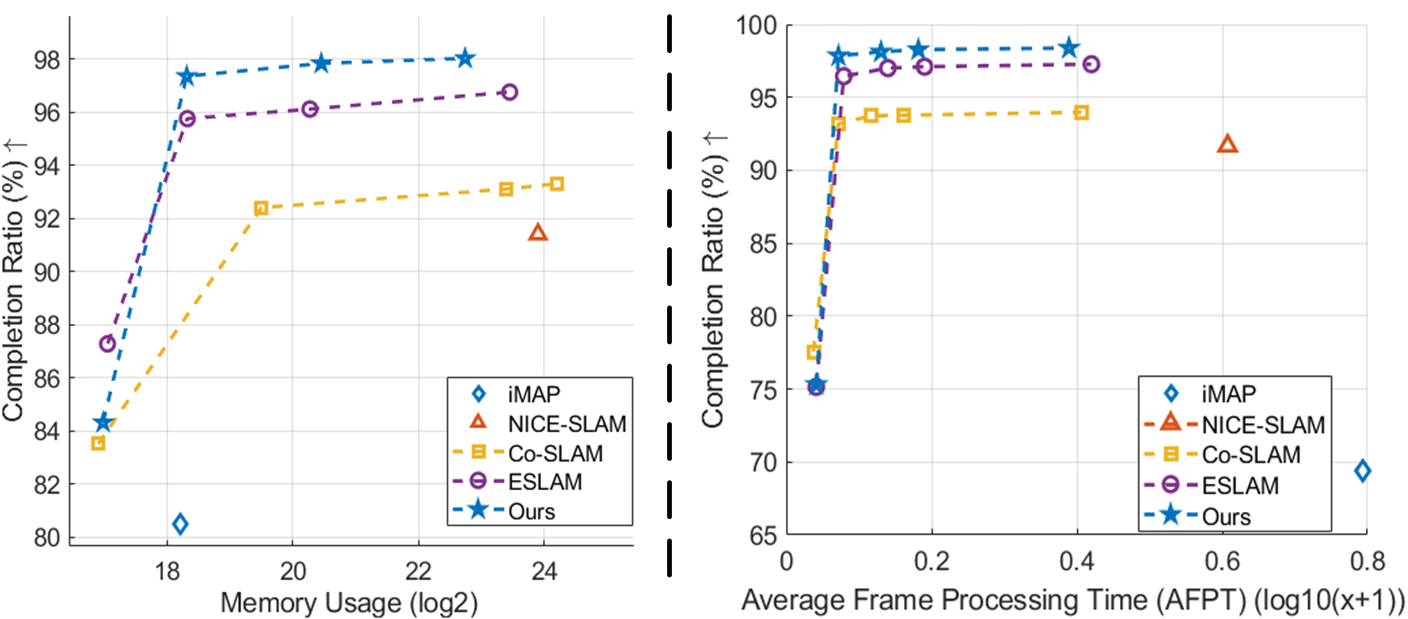}
   \vspace{-0.6cm}
    \caption{Completion ratio vs. model size and average time for PLGSLAM with other methods. Each model corresponds to a different hash-table or tri-planes size. }
   \label{fig:compare}
   \vspace{-0.6cm}
\end{figure}

\noindent \textbf{Metrics.}
We use Depth L1 (cm), Accuracy (cm), Completion (cm), and Completion ratio $(\%)$ to evaluate the reconstruction quality. Following NICE-SLAM and ESLAM\cite{niceslam,eslam}, we perform frustum and occlusion mesh culling that removes unobserved regions outside frustum and the noisy points within the
camera frustum but outside the target scene. However, this simple strategy removes too many meshes, leading to excessive holes and ineffective assessment of the reconstruction results. For the evaluation of camera tracking, we adopt ATE RMSE and Mean(cm).

\noindent \textbf{Implementation} We run PLGSLAM on a desktop PC with NVIDIA RTX 3090ti GPU. We employ feature planes
with a resolution of 24 cm for coarse tri-planes. We use 6 cm resolution for fine tri-planes. All feature planes have 32 channels, resulting in a 64-channel concatenated feature input for the decoders. The decoders are two-layer MLPs with 32 channels in the hidden layer.  For Replica \cite{replica},
we sample $N$ = 32 points for stratified sampling and
$N_{surface}$ = 8 points for importance sampling on each ray. And for ScanNet~\cite{scannet}, we set $N$ = 48 and $N_{surface}$ = 8. For further details of our implementation,
refer to the supplementary.
\vspace{-0.2cm}
\subsection{Experimental Results}
\vspace{-0.2cm}
\noindent \textbf{Replica dataset}. We evaluate on the same RGB-D sequences as ESLAM \cite{eslam} and Co-SLAM \cite{coslam}. We use this dataset to test our system performance in small room scenes (nearly $6.5m\times4.2m\times2.7m $). As shown in Tab. \ref{tab:replica}, our method achieves higher reconstruction and pose estimation accuracy. We show the qualitative results in Fig. \ref{fig:replica}. We
can see that ESLAM maintains more reconstruction details, but the results contain some artifacts. Co-SLAM achieves smooth completion in unobserved areas, but the accuracy of the reconstruction and pose estimation is relatively low. Our method successfully achieves consistent completion as well as high-fidelity reconstruction results.

\noindent\textbf{Scannet dataset.} We evaluate the camera tracking and reconstruction results of PLGSLAM on real-world large room sequences (nearly $7.5m\times6.6m\times2.7m $) from
ScanNet \cite{scannet} . We use the absolute trajectory error (ATE) as our metric. Tab. \ref{tab:scannet} shows that our method achieves better pose estimation and surface reconstruction results in comparison to NICE-SLAM \cite{niceslam}, ESLAM \cite{eslam}, and Co-SLAM \cite{coslam}. PLGSLAM exhibits superior scene representation capabilities and more accurate and robust tracking performance in large-scale indoor scenes. Fig. \cite{scannet} also shows PLGSLAM achieves better reconstruction quality
with smoother results and finer details.

\noindent\textbf{Apartment dataset.} We evaluate the surface reconstruction and camera tracking
accuracy of PLGSLAM on 
Apartment dataset (\textbf{nearly} $14.5m\times7.5m\times3.2m $).  Tab. \ref{tab:apartment} shows that quantitatively, our method achieves SOTA tracking results in comparison to Co-SLAM and ESLAM. These algorithms typically exhibit significant cumulative errors in large-scale indoor dataset scenarios. Fig. 1 also shows PLGSLAM achieves better reconstruction quality
with smoother results and finer details.

\subsection{Runtime analysis}

In this section, we analysis the speed and memory usage of our method compared with other SOTA methods in Replica datasets~\cite{replica} and ScanNet datasets~\cite{scannet}. We report the average frame pocessing time (FPT) and the memory growth rate in Tab.~\ref{tab:runtime}. The results indicate that our method is faster than previous methods and the model size does not grow cubically with the scene length. In Fig.~\ref{fig:compare}, we present the completion ratio under different model size and memory usage. We visualized the variation curve of the completion ratio by altering the size of the hash table/triplanes. 
\vspace{-0.2mm}
\subsection{Ablation Study}
\vspace{-0.2mm}
In this section, we conduct various experiments to verify the effectiveness of our method. 
Tab. \ref{tab:ablation} illustrates a quantitative evaluation with different settings.

 \noindent\textbf{Joint scene representation.} It is obvious that the joint scene representation (tri-planes with MLP) significantly improves our surface reconstruction accuracy. 

\noindent\textbf{Progressive scene representation.}
We replace our progressive scene representation and use a single network for the entire scene. We can observe that this method has a great influence on pose estimation and reconstruction metrics. This network significantly improves the capacity of scene geometry representation and enhances the robustness for local misestimation.

\noindent\textbf{Local-to-global bundle adjustment.}
We remove our local-to-global bundle adjustment in this experiment. Our full model leads to higher accuracy and better completion. The local-to-global BA can significantly reduce the growing cumulative error and improve the robustness and accuracy of the camera tracking.
\vspace{-0.3mm}
\section{Conclusion}
\vspace{-0.3mm}
In this paper, we propose a novel dense SLAM system, PLGSLAM, which achieve accurate surface reconstruction and pose estimation in large indoor scenes. Our progressive scene representation method enables our system to represent large-scale indoor scenes and long videos. The joint encoding method with the tri-planes and multi-layer perceptron further improves the accuracy of local scene representation. The local-to-global bundle adjustment method combines the traditional SLAM method with end-to-end pose estimation, which achieves robust and accurate camera tracking and mitigate the influence of cumulative error and pose drift. Our extensive experiments demonstrate the effectiveness and accuracy of our system in both scene reconstruction, depth estimation, and pose estimation. 

\textbf{Acknowledgement}
This work is supported by the National Key R\&D Program of China (Grant 2020YFC2007500), the National Natural Science Foundation of China (Grant U1813206), and the Science and Technology Commission of Shanghai Municipality (Grant 20DZ2220400).
Authors gratefully appreciate the contribution of Yanbo Wang from Shanghai Jiao Tong University, Hengyi Wang from University College London.
{\small
\bibliographystyle{ieee_fullname}
\bibliography{egbib}
}

\end{document}